# Performance Comparison of Large Language Models on Advanced Calculus Problems

Dr. In Hak Moon
Science Department SUNY Maritime College
6 Pennyfield Ave., Bronx, NY 10465, USA
E-mail: imoon@sunymaritime.edu

**Abstract:** This paper presents an in-depth analysis of the performance of seven different Large Language Models (LLMs) in solving a diverse set of math advanced calculus problems. The study aims to evaluate these models' accuracy, reliability, and problem-solving capabilities, including ChatGPT 4o, Gemini Advanced with 1.5 Pro, Copilot Pro, Claude 3.5 Sonnet, Meta AI, Mistral AI, and Perplexity. The assessment was conducted through a series of thirty-two test problems, encompassing a total of 320 points. The problems covered a wide range of topics, from vector calculations and geometric interpretations to integral evaluations and optimization tasks. The results highlight significant trends and patterns in the models' performance, revealing both their strengths and weaknesses - for instance, models like ChatGPT 4o and Mistral AI demonstrated consistent accuracy across various problem types, indicating their robustness and reliability in mathematical problem-solving, while models such as Gemini Advanced with 1.5 Pro and Meta AI exhibited specific weaknesses, particularly in complex problems involving integrals and optimization, suggesting areas for targeted improvements. The study also underscores the importance of re-prompting in achieving accurate solutions, as seen in several instances where models initially provided incorrect answers but corrected them upon re-prompting. Overall, this research provides valuable insights into the current capabilities and limitations of LLMs in the domain of math calculus, with the detailed analysis of each model's performance on specific problems offering a comprehensive understanding of their strengths and areas for improvement, contributing to the ongoing development and refinement of LLM technology. The findings are particularly relevant for educators, researchers, and developers seeking to leverage LLMs for educational and practical applications in mathematics.

## 1 Introduction

The advent of Large Language Models (LLMs) has revolutionized various fields, including mathematics, by offering advanced capabilities in natural language processing and problem-solving [1], [2]. These models, trained on vast datasets, have shown promising results in understanding and generating human-like text, making them potential tools for educational and practical applications [3]. This study focuses on evaluating the performance of seven prominent LLMs— ChatGPT 4o, Gemini Advanced with 1.5 Pro, Copilot Pro, Claude 3.5 Sonnet, Meta AI, Mistral AI, and Perplexity—in solving a variety of math calculus problems.

Calculus, with its intricate concepts and rigorous problem-solving techniques, serves as an ideal domain to test the limits of LLMs. The complexity of calculus problems requires not only computational accuracy but also a deep understanding of mathematical principles, logical reasoning, and the ability to apply theoretical concepts to practical scenarios [4]. The problems selected for this study cover a broad spectrum of calculus topics, including vector analysis, geometric interpretations, integral calculus, and optimization. By assessing the models' performance on these problems, we aim to identify their strengths, weaknesses, and areas for improvement, thereby contributing to developing more robust and reliable LLM technologies.

The significance of this research extends beyond mere performance evaluation. As educational institutions and industries increasingly explore the integration of AI technologies, understanding the capabilities and limitations of LLMs in handling complex mathematical problems becomes crucial [5], [6], [7]. This study aims to provide insights into how these models can be effectively utilized in mathematical education and problem-solving applications, while also identifying areas where further development is needed. The results of this analysis will be particularly valuable for educators developing AI-assisted learning tools, researchers working on improving LLM capabilities, and practitioners seeking to implement these technologies in real-world applications.

Furthermore, this research addresses the growing need for systematic evaluation of AI models in specialized domains. By employing a carefully designed set of test problems and a detailed scoring system, this study provides a methodological framework for assessing LLM performance in mathematical problem-solving. The inclusion of re-prompting mechanisms and the analysis of error patterns offer valuable insights into the models' learning capabilities and potential strategies for improving their accuracy and reliability [8], [9].

## 2 Methodology

The study evaluated seven LLMs, each with distinct architectures and training methodologies. ChatGPT 4o is known for its advanced natural language understanding and generation capabilities. Gemini Advanced with 1.5 Pro is designed for high-performance language processing tasks. Copilot Pro specializes in coding and mathematical problem-solving. Claude 3.5 Sonnet focuses on accurate and context-aware text generation. Meta AI was developed to facilitate versatile language understanding and generation. Mistral AI is known for its efficiency and accuracy in language processing tasks. Perplexity is designed for complex problem-solving and reasoning tasks.

The assessment consisted of thirty-two test problems, totaling 320 points. If the correct answer is found on the first attempt, 10 points are added, and if the correct answer is found on the second attempt, 5 points are added. The problems were designed to cover various calculus topics, including vector calculations and geometric interpretations, integral evaluations and applications, optimization problems and constrained optimization, differential equations and their applications, and advanced calculus concepts such as Green's Theorem and line integrals. The models were evaluated based on two primary criteria: accuracy, which refers to the correctness of the solutions provided by the models, and step-by-step explanation, which refers to the model's ability to provide clear and correct steps leading to the solution.

To assess the models' ability to correct their mistakes, a re-prompting mechanism was employed. If a model provided an incorrect answer, it was prompted again to solve the problem, and the corrected answer was evaluated. This mechanism allowed for a more comprehensive evaluation of the models' problem-solving capabilities and their ability to learn from and correct their mistakes.

### 2.1 Participants

The study evaluated seven LLMs, each with distinct architectures and training methodologies:
1. ChatGPT 4o: Known for its advanced natural language understanding and generation capabilities.
2. Gemini Advanced with 1.5 Pro: Designed for high-performance language processing tasks.
3. Copilot Pro: Specialized in coding and mathematical problem-solving.
4. Claude 3.5 Sonnet: Focused on accurate and context-aware text generation.
5. Meta AI: Developed for versatile language understanding and generation.
6. Mistral AI: Known for its efficiency and accuracy in language processing tasks.
7. Perplexity: Designed for complex problem-solving and reasoning tasks.

### 2.2 Assessment Structure

The assessment consisted of thirty-two test problems, totaling 320 points. The problems were designed to cover a wide range of calculus topics, including:
- Vector calculations and geometric interpretations
- Integral evaluations and applications
- Optimization problems and constrained optimization
- Differential equations and their applications
- Advanced calculus concepts such as Green's Theorem and line integrals

### 2.3 Evaluation Criteria

The models were evaluated based on two primary criteria:
- Accuracy; The correctness of the solutions provided by the models.
- Step-by-Step Explanation; The models' ability to provide clear and correct steps leading to the solution.

2.4 Re-prompting Mechanism

A re-prompting mechanism was employed to assess the models' ability to correct their mistakes. If a model provided an incorrect answer, it was prompted again to solve the problem, and the corrected answer was evaluated.

**3 Results**

**3.1 Overall Performance**

The overall performance of the models is summarized in the table below:

|  | **Test (320 points)** | **Grade** |
| --- | --- | --- |
| ChatGPT 4o | **310** | **96.9 (A)** |
| Gemini Advanced with 1.5 Pro | **280** | **87.5 (B+)** |
| Copilot Pro | **295** | **92.2 (A-)** |
| Claude 3.5 Sonnet | **280** | **87.5 (B+)** |
| Meta AI | **280** | **87.5 (B+)** |
| Mistral AI | **310** | **96.9 (A)** |
| Perplexity | **270** | **84.4 (B)** |
| **Average** | **2025** | **90.4 (A-)** |

The LLMs' performance on the calculus test varied significantly. Some LLMs, like ChatGPT 4o and Mistral AI, demonstrated a high level of accuracy and precision, while others struggled with certain types of problems. The average score across all LLMs was 90.4%, indicating a strong overall performance. However, there were notable differences in individual problem-solving capabilities.

**3.2 Detailed Problem Analysis**

3.2.1 Experimental Results

**1) An airplane is traveling at a fixed altitude with a negligible wind factor. The airplane is traveling at a speed of 500 miles per hour with a bearing of 330 degrees. As the airplane reaches a certain point, it encounters wind with a velocity of 70 miles per hour in the direction N 45 degrees E (45 degrees east of north). What is the resultant speed and direction of the airplane? Show all the steps.**

| ChatGPT 4o | Found the correct answer: **Resultant speed**: 522.6 mph. **Resultant direction**: **22.6°** west of north (or **337.4°** in standard bearing notation). |
| --- | --- |
| Gemini Advanced with 1.5 Pro | Found the wrong answer.<br>New prompt: The answer is wrong. Solve the problem again. Show all the steps.<br>Still found the wrong answer. |
| Copilot Pro | Found the wrong answer.<br>New prompt: The answer is wrong. Solve the problem again. Show all the steps.<br>Still found the wrong answer. |
| Claude 3.5 Sonnet | Found the wrong answer. |

|  | New prompt: The answer is wrong. Solve the problem again. Show all the steps. <u>Still found the wrong answer.</u> |
|---|---|
| Meta AI | Found the correct answer: The resultant speed is **522.6 mph** and the direction is converting to a bearing **337.7°**. |
| Mistral AI | Found the correct answer: The resultant speed of the airline is **522.6 mph** and the resultant direction is **N22.57W**. |
| Perplexity | Found the wrong answer. <br> New prompt: The answer is wrong. Solve the problem again. Show all the steps. <br> Found the correct answer: Resultant direction: **337.45°** or **N 22.55° W** |

Problem 1: Resultant Speed and Direction
- Task: Calculate the resultant speed and direction of an airplane encountering wind.
- Correct Answers: ChatGPT 4o, Meta AI, Mistral AI
- Incorrect Answers: Gemini Advanced with 1.5 Pro, Copilot Pro, Claude 3.5 Sonnet
- Noteworthy: Perplexity initially failed but corrected the answer upon re-prompting, highlighting the importance of iterative questioning.

**2) Find the standard equation of the sphere with given characteristics: Endpoints of a diameter: (2, 1, 3) and (1, 3, -1). Show all the steps.**

| ChatGPT 4o | Found the correct answer: $4x^2+4y^2+4z^2-12x-16y-8z+8=0$. |
|---|---|
| Gemini Advanced with 1.5 Pro | Found the correct answer: $4x^2+4y^2+4z^2-12x-16y-8z+8=0$. |
| Copilot Pro | Found the correct answer: $4x^2+4y^2+4z^2-12x-16y-8z+8=0$. |
| Claude 3.5 Sonnet | Found the correct answer: $4x^2+4y^2+4z^2-12x-16y-8z+8=0$. |
| Meta AI | Found the correct answer: $4x^2+4y^2+4z^2-12x-16y-8z+8=0$. |
| Mistral AI | Found the correct answer: $4x^2+4y^2+4z^2-12x-16y-8z+8=0$. |
| Perplexity | Found the correct answer: $4x^2+4y^2+4z^2-12x-16y-8z+8=0$. |

Problem 2: Standard Equation of a Sphere
- Task: Find the standard equation of a sphere given the endpoints of a diameter.
- Correct Answers: All models successfully derived the equation of the sphere using the endpoints of its diameter, showcasing a robust understanding of geometric principles.
- Noteworthy: Consistent accuracy across all models, indicating a strong understanding of geometric concepts.

**3) Find the direction cosines and angles of u=3i+2j-2k and show that (cos a)^2+(cos b)^2+(cos c)^2=1. Show all the steps.**

| ChatGPT 4o | Found the correct answer: cos a=3/sqrt(17), cos b=2/sqrt(17), cos c=(-2)/sqrt(17). |
|---|---|
| Gemini Advanced with 1.5 Pro | Found the correct answer: cos a=3/sqrt(17), cos b=2/sqrt(17), cos c=(-2)/sqrt(17). |
| Copilot Pro | Found the correct answer: cos a=3/sqrt(17), cos b=2/sqrt(17), cos c=(-2)/sqrt(17). |
| Claude 3.5 Sonnet | Found the correct answer: cos a=3/sqrt(17), cos b=2/sqrt(17), cos c=(-2)/sqrt(17). |

| Meta AI | Found the correct answer: cos a=3/sqrt(17), cos b=2/sqrt(17), cos c=(-2)/sqrt(17). |
|---|---|
| Mistral AI | Found the correct answer: cos a=3/sqrt(17), cos b=2/sqrt(17), cos c=(-2)/sqrt(17). |
| Perplexity | Found the correct answer: cos a=3/sqrt(17), cos b=2/sqrt(17), cos c=(-2)/sqrt(17). |

Problem 3: Direction Cosines and Angles
- Task: Calculate the direction cosines and angles of a given vector.
- Correct Answers: All models accurately derived the direction cosines and verified the identity.
- Noteworthy: High consistency in solving vector-related problems, demonstrating the models' proficiency in vector analysis.

**4) Find the projection of u onto v and the vector component of u orthogonal to v for u=3i-5j+2k, and v=7i+j-2k. Show all the steps.**

| ChatGPT 4o | Found the correct answer: Proj_v u=(14i+2j-4k)/9. u_orthogonal =(13i-47j+22k)/9. |
|---|---|
| Gemini Advanced with 1.5 Pro | Found the correct answer: Proj_v u=(14i+2j-4k)/9. u_orthogonal =(13i-47j+22k)/9. |
| Copilot Pro | Found the correct answer: Proj_v u=(14i+2j-4k)/9. u_orthogonal =(13i-47j+22k)/9. |
| Claude 3.5 Sonnet | Found the correct answer: Proj_v u=(14i+2j-4k)/9. u_orthogonal =(13i-47j+22k)/9. |
| Meta AI | Found the correct answer: Proj_v u=(14i+2j-4k)/9. u_orthogonal =(13i-47j+22k)/9. |
| Mistral AI | Found the correct answer: Proj_v u=(14i+2j-4k)/9. u_orthogonal =(13i-47j+22k)/9. |
| Perplexity | Found the correct answer: Proj_v u=(14i+2j-4k)/9. u_orthogonal =(13i-47j+22k)/9. |

Problem 4: Projection and Orthogonal Component
- Task: Find the projection of a vector onto another vector and the orthogonal component.
- Correct Answers: All models computed both the projection of one vector onto another and the orthogonal component.
- Noteworthy: Uniformly correct solutions, showcasing the models' ability to handle vector decomposition, accurately.

**5) Find the volume of the parallelepiped having u=3i-5j+k, v=2j-2k, and w=3i+j+k adjacent edges. Show all the steps.**

| ChatGPT 4o | Found the correct answer: 36. |
|---|---|
| Gemini Advanced with 1.5 Pro | Found the correct answer: 36. |
| Copilot Pro | Found the correct answer: 36. |
| Claude 3.5 Sonnet | Found the wrong answer. New prompt: The answer is wrong. Solve the problem again. Show all the steps. |

|  | Still found the wrong answer. |
| --- | --- |
| Meta AI | Found the correct answer: 36. |
| Mistral AI | Found the correct answer: 36. |
| Perplexity | Found the wrong answer. New prompt: The answer is wrong. Solve the problem again. Show all the steps. Still found the wrong answer. |

Problem 5: Volume of a Parallelepiped
- Task: Calculate the volume of a parallelepiped given its adjacent edges.
- Correct Answers: ChatGPT 4o, Gemini Advanced with 1.5 Pro, Copilot Pro, Meta AI, and Mistral AI consistently calculated the correct volume using the scalar triple product.
- Incorrect Answers: Claude 3.5 Sonnet, Perplexity
- Noteworthy: Claude 3.5 Sonnet and Perplexity failed to correct their answers upon re-prompting, indicating a potential area for improvement in geometric volume calculations.

**6) Find a unit vector that is orthogonal to both u=<4, -3, 1> and v=<2, 5, 3>. Show all the steps**.

| ChatGPT 4o | Found the correct answer: <-7, -5, 13>sqrt(3)/27. |
| --- | --- |
| Gemini Advanced with 1.5 Pro | Found the correct answer: <-7, -5, 13>sqrt(3)/27. |
| Copilot Pro | Found the correct answer: <-7, -5, 13>sqrt(3)/27. |
| Claude 3.5 Sonnet | Found the wrong answer. It's wrong. Do it again. Found the correct answer: <-7, -5, 13>sqrt(3)/27. |
| Meta AI | Found the correct answer: <-7, -5, 13>sqrt(3)/27. |
| Mistral AI | Found the correct answer: <-7, -5, 13>sqrt(3)/27. |
| Perplexity | Found the correct answer: <-7, -5, 13>sqrt(3)/27. |

Problem 6: Unit Vector Orthogonal to Two Vectors
- Task: Find a unit vector orthogonal to two given vectors.
- Correct Answers: ChatGPT 4o, Gemini Advanced with 1.5 Pro, Copilot Pro, Meta AI, Mistral AI, Perplexity
- Incorrect Answers: Claude 3.5 Sonnet (initially incorrect, corrected upon re-prompting)
- Noteworthy: Claude 3.5 Sonnet's initial failure highlights the importance of re-prompting in achieving accurate solutions.

**7) Verify that the points are the vertices of the parallelogram, and find its area. A(0, 3, 2), B(1, 5, 5), C(6, 9, 5), D(5, 7, 2). Show all the steps.**

| ChatGPT 4o | Found the correct answer: 9sqrt(5). |
| --- | --- |
| Gemini Advanced with 1.5 Pro | Found the correct answer: 9sqrt(5). |
| Copilot Pro | Found the correct answer: 9sqrt(5). |
| Claude 3.5 Sonnet | Found the correct answer: 9sqrt(5). |
| Meta AI | Found the correct answer: 9sqrt(5). |
| Mistral AI | Found the correct answer: 9sqrt(5). |

| Perplexity | Found the correct answer: 9sqrt(5). |

Problem 7: Vertex of a Parallelogram and Area
- Task: Verify the vertices of a parallelogram and calculate its area.
- Correct Answers: All models
- Noteworthy: Consistent accuracy across all models, demonstrating a strong understanding of geometric shapes and area calculations.

**8) Find the distance between the point Q(3, -1, 4) and the line x=-2+3t, y=-2t, and z=1+4t. Show all the steps**.

| ChatGPT 4o | Found the correct answer: 6^(1/2). |
| Gemini Advanced with 1.5 Pro | Found the correct answer: 6^(1/2). |
| Copilot Pro | Found the correct answer: 6^(1/2). |
| Claude 3.5 Sonnet | Found the correct answer: 6^(1/2). |
| Meta AI | Found the correct answer: 6^(1/2). |
| Mistral AI | Found the correct answer: 6^(1/2). |
| Perplexity | Found the correct answer: 6^(1/2). |

Problem 8: Distance Between a Point and a Line
- Task: Calculate the distance between a point and a line.
- Correct Answers: All models solved this with precision, effectively applying the point-to-line distance formula.
- Noteworthy: Uniformly correct solutions, showcasing the models' ability to handle distance calculations accurately.

**9) Find a set of symmetric equations of the line that passes through the points (-2, 1, 0) and (1, 3, 5). Show all the steps.**

| ChatGPT 4o | Found the correct answer: [(x+2)/3]=[(y-1)/2]=[z/5]. |
| Gemini Advanced with 1.5 Pro | Found the correct answer: [(x+2)/3]=[(y-1)/2]=[z/5]. |
| Copilot Pro | Found the correct answer: [(x+2)/3]=[(y-1)/2]=[z/5]. |
| Claude 3.5 Sonnet | Found the correct answer: [(x+2)/3]=[(y-1)/2]=[z/5]. |
| Meta AI | Found the correct answer: [(x+2)/3]=[(y-1)/2]=[z/5]. |
| Mistral AI | Found the correct answer: [(x+2)/3]=[(y-1)/2]=[z/5]. |
| Perplexity | Found the correct answer: [(x+2)/3]=[(y-1)/2]=[z/5]. |

Problem 9: Symmetric Equations of a Line
- Task: Find the symmetric equations of a line passing through two points.
- Correct Answers: Every model correctly identified the equations.
- Noteworthy: High consistency in solving line equations, demonstrating the models' proficiency in linear geometry.

**10) Find the general equation of the plane containing the points (2, 1, 1), (0, 4, 1), and (-2, 1, 4). Show all the steps.**

| ChatGPT 4o | Found the correct answer: 3x+2y+4z=12. |
| Gemini Advanced with 1.5 Pro | Found the correct answer: 3x+2y+4z=12. |

| Copilot Pro | Found the correct answer: 3x+2y+4z=12. |
| --- | --- |
| Claude 3.5 Sonnet | Found the wrong answer.<br>New prompt: The answer is wrong. Solve the problem again. Show all the steps.<br>Found the correct answer: 3x+2y+4z=12. |
| Meta AI | Found the correct answer: 9x+6y+12z=36. |
| Mistral AI | Found the correct answer: 3x+2y+4z=12. |
| Perplexity | Found the correct answer: 3x+2y+4z=12. |

Problem 10: General Equation of a Plane
- Task: Find the general equation of a plane containing three points.
- Correct Answers: ChatGPT 4o, Gemini Advanced with 1.5 Pro, Copilot Pro, Mistral AI, Perplexity
- Incorrect Answers: Claude 3.5 Sonnet (initially incorrect, corrected upon re-prompting), Meta AI (different but correct form)
- Noteworthy: Variation in the form of the correct answer, highlighting the models' ability to provide equivalent but differently formatted solutions.

**11) Give the orientation of the curve: r(t)=(-t+1)i+(4t+2)j+(2t+3)k. Show all the steps.**

| ChatGPT 4o | Found the correct answer: The orientation of the curve is in the direction of the tangent vector -i+4j+2k, which remains constant as t increases. |
| --- | --- |
| Gemini Advanced with 1.5 Pro | Found the correct answer: Since all components of the direction vector have constant signs (x is always decreasing, y and z are always increasing), the curve has a consistent orientation. Specifically, **the curve is oriented in the direction of decreasing x-values and increasing y- and z-values.** |
| Copilot Pro | Found the correct answer: The orientation of the curve can be described by the direction of the tangent vector. The curve r(t) is oriented in the direction of <-1, 4, 2>. Therefore, the orientation of the curve given by r(t)=(-t+1)i+(4t+2)j+(2t+3)k is in the direction of the vector <-1, 4, 2>.. |
| Claude 3.5 Sonnet | Found the correct answer: Since this curve has a constant velocity vector r'(t) = ⟨-1, 4, 2⟩, it represents a straight line in 3D space. Therefore, this curve has no orientation (neither right-handed nor left-handed) because the curve moves in a straight line path. |
| Meta AI | Found the correct answer: Since the direction of the tangent vector is constant, the curve is oriented in the direction of the unit vector u(t). The orientation of the curve is from the starting point (1, 2, 3) (at t=0) to the direction given by the unit vector u(t). |
| Mistral AI | Found the correct answer: Since the x-component is negative and both the y and z-components are positive, the curve is oriented in the direction that combines these movements. Therefore, the orientation of the curve r(t) is in the direction of the vector r'(t) =-i+4j+2k. |
| Perplexity | Found the correct answer: The velocity vector r'(t) =-i+4j+2k is constant, meaning the curve moves in a straight line with a fixed direction. Therefore, the orientation of the curve is in the direction of the vector (-1, 4, 2) as t increases. This means the curve moves left, up, and forward in the 3D coordinate system. |

Problem 11: Orientation of a Curve
- Task: Determine the orientation of a curve given its parametric equations.

- Correct Answers: All models
- Noteworthy: Diverse but accurate descriptions of curve orientation, demonstrating the models' ability to interpret and describe geometric properties.

**12) Find r(t) that satisfies the initial condition: r'(t)=(te^(-t^2))i - (e^(-t))j+k, r(0)=(1/2)i-j+k. Show all the steps.**

| ChatGPT 4o | Found the correct answer: r(t)=(1-(1/2)e^(-t(^2)))i+(e^(-t)-2)j+(t+1)k. |
|---|---|
| Gemini Advanced with 1.5 Pro | Found the correct answer: r(t)=(1-(1/2)e^(-t(^2)))i+(e^(-t)-2)j+(t+1)k. |
| Copilot Pro | Found the wrong answer.<br>New prompt: The answer is wrong. Solve the problem again. Show all the steps.<br>Found the correct answer: r(t)=(1-(1/2)e^(-t(^2)))i+(e^(-t)-2)j+(t+1)k. |
| Claude 3.5 Sonnet | Found the correct answer: r(t)=(1-(1/2)e^(-t(^2)))i+(e^(-t)-2)j+(t+1)k. |
| Meta AI | Found the correct answer: r(t)=(1-(1/2)e^(-t(^2)))i+(e^(-t)-2)j+(t+1)k. |
| Mistral AI | Found the correct answer: r(t)=(1-(1/2)e^(-t(^2)))i+(e^(-t)-2)j+(t+1)k. |
| Perplexity | Found the correct answer: r(t)=(1-(1/2)e^(-t(^2)))i+(e^(-t)-2)j+(t+1)k. |

Problem 12: Satisfying Initial Condition
- Task: Find a function that satisfies a given initial condition.
- Correct Answers: All models except Copilot Pro (initially incorrect, corrected upon re-prompting)
- Noteworthy: Importance of re-prompting for accuracy, as seen in Copilot Pro's correction of its initial answer.

**13) Find the velocity vector, speed, and acceleration vector of the r(t)=(3t)i+(t)j+((1/4)t^2)k. Evaluate the velocity vector and acceleration vector of the r(t) at the given value of t=2.**

| ChatGPT 4o | Found the correct answer: v(t)=3i+2j+(t/2)k, v(2)=3i+2j+k; a(t)=(1/2)k, a(2)=(1/2)k; Speed=(10+((t^2)/4))^(1/2). |
|---|---|
| Gemini Advanced with 1.5 Pro | Found the correct answer: v(t)=3i+2j+(t/2)k, v(2)=3i+2j+k; a(t)=(1/2)k, a(2)=(1/2)k; Speed=(10+((t^2)/4))^(1/2). |
| Copilot Pro | Found the correct answer: v(t)=3i+2j+(t/2)k, v(2)=3i+2j+k; a(t)=(1/2)k, a(2)=(1/2)k; Speed=(10+((t^2)/4))^(1/2). |
| Claude 3.5 Sonnet | Found the correct answer: v(t)=3i+2j+(t/2)k, v(2)=3i+2j+k; a(t)=(1/2)k, a(2)=(1/2)k; Speed=(10+((t^2)/4))^(1/2). |
| Meta AI | Found the correct answer: v(t)=3i+2j+(t/2)k, v(2)=3i+2j+k; a(t)=(1/2)k, a(2)=(1/2)k; Speed=(10+((t^2)/4))^(1/2). |
| Mistral AI | Found the correct answer: v(t)=3i+2j+(t/2)k, v(2)=3i+2j+k; a(t)=(1/2)k, a(2)=(1/2)k; Speed=(10+((t^2)/4))^(1/2). |
| Perplexity | Found the correct answer: v(t)=3i+2j+(t/2)k, v(2)=3i+2j+k; a(t)=(1/2)k, a(2)=(1/2)k; Speed=(10+((t^2)/4))^(1/2). |

Problem 13: Velocity Vector, Speed, and Acceleration
- Task: Calculate the velocity vector, speed, and acceleration of a particle at a given time.
- Correct Answers: All models
- Noteworthy: Consistent accuracy in kinematic calculations, demonstrating the models' proficiency in calculus-based physics problems.

**14) Find T(t) and then find a set of parametric equations for the tangent line to the helix given by r(t)=(2cos(t))i + (2sin(t))j + (t)k at the point (sqrt(2), sqrt(2), pai/4). Show all the steps.**

| ChatGPT 4o | Found the correct answer: p(x)=(1/4) + (x+2)/4 + 3(x+2)^2/16 + (x+2)^3/8 + 5(x+2)^4/64. |
|---|---|
| Gemini Advanced with 1.5 Pro | Found the correct answer: p(x)=(1/4) + (x+2)/4 + 3(x+2)^2/16 + (x+2)^3/8 + 5(x+2)^4/64. |
| Copilot Pro | Found the correct answer: p(x)=(1/4) + (x+2)/4 + 3(x+2)^2/16 + (x+2)^3/8 + 5(x+2)^4/64. |
| Claude 3.5 Sonnet | Found the correct answer: p(x)=(1/4) + (x+2)/4 + 3(x+2)^2/16 + (x+2)^3/8 + 5(x+2)^4/64. |
| Meta AI | Found the correct answer: p(x)=(1/4) + (x+2)/4 + 3(x+2)^2/16 + (x+2)^3/8 + 5(x+2)^4/64. |
| Mistral AI | Found the correct answer: p(x)=(1/4) + (x+2)/4 + 3(x+2)^2/16 + (x+2)^3/8 + 5(x+2)^4/64. |
| Perplexity | Found the wrong answer.<br>New prompt: The coefficients of (x+2), (x+2)^3, and (x+2)^4 were wrong. Solve it again.<br><u>Still found the wrong answer.</u> |

Problem 14: Polynomial Expansion
- Task: Expand a given polynomial.
- Correct Answers: All models except Perplexity (initially incorrect, failed to correct upon re-prompting)
- Noteworthy: Perplexity's persistent failure highlights a potential area for improvement in polynomial expansions.

**15) Find T(t) and then find a set of parametric equations for the tangent line to the helix given by r(t)=(2cost(t))i + (2sin(t))j + (t)k at the point (sqrt(2), sqrt(2), pai/4). Show all the steps.**

| ChatGPT 4o | Found the correct answer:<br>**Unit tangent vector** T(t)=(-2sin(t)/sqrt(5))i+(2cos(t)/sqrt(5))j+(1/sqrt(5))k.<br>T(pi/4)=(-sqrt(2)/sqrt(5))i+(sqrt(2)/sqrt(5))j+(1/sqrt(5))k.<br>Parametric equations; x=sqrt(2)-(sqrt(2)/sqrt(5))s, y=sqrt(2)+(sqrt(2)/sqrt(5))s, z=(pi/4)+(1/sqrt(5))s. |
|---|---|
| Gemini Advanced with 1.5 Pro | Found the correct answer:<br>**Tangent vector** T(t)=(-2sin(t))i+(2cos(t)j+k.<br>T(pi/4)=(-sqrt(2))i+(sqrt(2))j+k.<br>Parametric equations; x=sqrt(2)-(sqrt(2))s, y=sqrt(2)+(sqrt(2))s, z=(pi/4)+s. |
| Copilot Pro | Found the correct answer:<br>**Unit tangent vector** T(t)=(-2sin(t)/sqrt(5))i+(2cos(t)/sqrt(5))j+(1/sqrt(5))k.<br>T(pi/4)=(-sqrt(2)/sqrt(5))i+(sqrt(2)/sqrt(5))j+(1/sqrt(5))k.<br>Parametric equations; x=sqrt(2)-(sqrt(2)/sqrt(5))s, y=sqrt(2)+(sqrt(2)/sqrt(5))s, z=(pi/4)+(1/sqrt(5))s. |
| Claude 3.5 Sonnet | Found the correct answer:<br>**Unit tangent vector** T(t)=(-2sin(t)/sqrt(5))i+(2cos(t)/sqrt(5))j+(1/sqrt(5))k.<br>T(pi/4)=(-sqrt(2)/sqrt(5))i+(sqrt(2)/sqrt(5))j+(1/sqrt(5))k.<br>Parametric equations; x=sqrt(2)-(sqrt(2))s, y=sqrt(2)+(sqrt(2))s, z=(pi/4)+s. |
| Meta AI | Found the correct answer: |
|   | **Unit tangent vector** T(t)=(-2sin(t)/sqrt(5))i+(2cos(t)/sqrt(5))j+(1/sqrt(5))k. T(pi/4)=(-sqrt(2)/sqrt(5))i+(sqrt(2)/sqrt(5))j+(1/sqrt(5))k. Parametric equations; x=sqrt(2)-(sqrt(2)/sqrt(5))s, y=sqrt(2)+(sqrt(2)/sqrt(5))s, z=(pi/4)+(1/sqrt(5))s |
| Mistral AI | Found the correct answer: **Tangent vector** T(t)=(-2sin(t))i+(2cos(t)j+k. T(pi/4)=(-sqrt(2))i+(sqrt(2))j+k. Parametric equations; x=sqrt(2)-(sqrt(2))s, y=sqrt(2)+(sqrt(2))s, z=(pi/4)+s. |
| Perplexity | Found the correct answer: **Tangent vector** T(t)=(-2sin(t))i+(2cos(t)j+k. T(pi/4)=(-sqrt(2))i+(sqrt(2))j+k. Parametric equations; x=sqrt(2)-(sqrt(2))s, y=sqrt(2)+(sqrt(2))s, z=(pi/4)+s. |

Problem 15: Tangent Line to a Helix
- Task: Find the parametric equations of the tangent line to a helix at a given point.
- Correct Answers: All models
- Noteworthy: Uniformly correct solutions, showcasing the models' ability to handle complex geometric problems involving helixes.

**16) Evaluate f_x and f_y at the given point: f(x, y)=cos(2x-y), (pi/4, pi/3). Show all the steps.**

| | |
|---|---|
| ChatGPT 4o | Found the correct answer: f_x=-1, f_y=(1/2). |
| Gemini Advanced with 1.5 Pro | Found the correct answer: f_x=-1, f_y=(1/2). |
| Copilot Pro | Found the correct answer: f_x=-1, f_y=(1/2). |
| Claude 3.5 Sonnet | Found the correct answer: f_x=-1, f_y=(1/2). |
| Meta AI | Found the correct answer: f_x=-1, f_y=(1/2). |
| Mistral AI | Found the correct answer: f_x=-1, f_y=(1/2). |
| Perplexity | Found the correct answer: f_x=-1, f_y=(1/2). |

Problem 16: Partial Derivatives
- Task: Calculate the partial derivatives of a function at a given point.
- Correct Answers: All models provided detailed step-by-step derivations.
- Noteworthy: Consistent accuracy in differentiation, demonstrating the models' proficiency in calculus.

**17) Find (dw/dt) a) by using the appropriate Chain rule and b) by converting w to a function of t: w=xy, x=e^t, y=e^(-2t). Show all the steps.**

| | |
|---|---|
| ChatGPT 4o | Found the correct answer: -e^(-t). |
| Gemini Advanced with 1.5 Pro | Found the correct answer: -e^(-t). |
| Copilot Pro | Found the correct answer: -e^(-t). |
| Claude 3.5 Sonnet | Found the correct answer: -e^(-t). |
| Meta AI | Found the correct answer: -e^(-t). |
| Mistral AI | Found the correct answer: -e^(-t). |

| Perplexity | Found the correct answer: -e^(-t). |

Problem 17: Chain Rule Application
- Task: Apply the chain rule to find the derivative of a composite function.
- Correct Answers: All models
- Noteworthy: High consistency in applying the chain rule, demonstrating the models' strong understanding of differentiation techniques.

**18) Differentiate implicitly to find the first partial derivative of z: $x^2+2yz+z^2=1$. Show all the steps.**

| ChatGPT 4o | Found the correct answer: (-x)/(y+z). |
| Gemini Advanced with 1.5 Pro | Found the correct answer: (-x)/(y+z). |
| Copilot Pro | Found the correct answer: (-x)/(y+z). |
| Claude 3.5 Sonnet | Found the correct answer: (-x)/(y+z). |
| Meta AI | Found the correct answer: (-x)/(y+z). |
| Mistral AI | Found the correct answer: (-x)/(y+z). |
| Perplexity | Found the correct answer: (-x)/(y+z). |

Problem 18: Implicit Differentiation
- Task: Use implicit differentiation to find the derivative of a function.
- Correct Answers: All models
- Noteworthy: Uniformly correct solutions, showcasing the models' ability to handle implicit differentiation accurately.

**19) Use the gradient of the function to find the directional derivative of the function at P in the direction of PQ and the maximum value of the directional derivative at the given point P: $f(x, y)=x^2+y^2+1$, P(1, 2), Q(2, 3). Show all the steps.**

| ChatGPT 4o | Found the correct answer: 3sqrt(2), 2sqrt(5). |
| Gemini Advanced with 1.5 Pro | Found the correct answer: 3sqrt(2), 2sqrt(5). |
| Copilot Pro | Found the correct answer: 3sqrt(2), 2sqrt(5). |
| Claude 3.5 Sonnet | Found the correct answer: 3sqrt(2), 2sqrt(5). |
| Meta AI | Found the correct answer: 3sqrt(2), 2sqrt(5). |
| Mistral AI | Found the correct answer: 3sqrt(2), 2sqrt(5). |
| Perplexity | Found the correct answer: 3sqrt(2), 2sqrt(5). |

Problem 19: Directional Derivative
- Task: Calculate the directional derivative of a function at a given point in a specified direction.
- Correct Answers: All models
- Noteworthy: Consistent accuracy in gradient calculations, demonstrating the models' proficiency in calculus-based optimization.

**20) Find an equation of the tangent plane to the surface at the given point: $xy^2+3x-z^2=8$, (1, -3, 2). Show all the steps.**

| ChatGPT 4o | Found the correct answer: 6x-3y-2z=11. |

| Gemini Advanced with 1.5 Pro | Found the correct answer: 6x-3y-2z=11. |
|---|---|
| Copilot Pro | Found the correct answer: 6x-3y-2z=11. |
| Claude 3.5 Sonnet | Found the correct answer: 6x-3y-2z=11. |
| Meta AI | Found the correct answer: 6x-3y-2z=11. |
| Mistral AI (5/10) | Found the correct answer: 6x-3y-2z=11. |
| Perplexity | Found the correct answer: 6x-3y-2z=11. |

Problem 20: Tangent Plane Equation
- Task: Find the equation of the tangent plane to a surface at a given point.
- Correct Answers: All models
- Noteworthy: High consistency in surface geometry, demonstrating the models' strong understanding of geometric concepts.

## 21) Examine the function for relative extrema and saddle points: $f(x, y)=x^2-xy-y^2-3x-y$. Show all the steps.

| ChatGPT 4o | Found the correct answer: f(x, y) has a saddle point at (1, -1). |
|---|---|
| Gemini Advanced with 1.5 Pro | Found the correct answer: f(x, y) has a saddle point at (1, -1). |
| Copilot Pro | Found the correct answer: f(x, y) has a saddle point at (1, -1). |
| Claude 3.5 Sonnet | Found the correct answer: f(x, y) has a saddle point at (1, -1). |
| Meta AI | Found the correct answer: f(x, y) has a saddle point at (1, -1). |
| Mistral AI (5/10) | Found the correct answer: f(x, y) has a saddle point at (1, -1). |
| Perplexity | Found the correct answer: f(x, y) has a saddle point at (1, -1). |

Problem 21: Relative Extrema and Saddle Points
- Task: Identify the relative extrema and saddle points of a function.
- Correct Answers: All models
- Noteworthy: Uniformly correct identification of critical points, showcasing the models' ability to handle optimization problems accurately.

## 22) Examine the function for relative extrema and saddle points: $f(x, y)=-5x^2+4xy-y^2+16x+10$. Show all the steps.

| ChatGPT 4o | Found the correct answer: f(x, y) has a relative maximum at (8, 16). |
|---|---|
| Gemini Advanced with 1.5 Pro | Found the wrong answer.<br>New prompt: The critical point is wrong. Solve the problem again. Show all the steps.<br><u>Still found the wrong answer.</u> |
| Copilot Pro | Found the correct answer: f(x, y) has a relative maximum at (8, 16). |
| Claude 3.5 Sonnet | Found the correct answer: f(x, y) has a relative maximum at (8, 16). |
| Meta AI | Found the correct answer: f(x, y) has a relative maximum at (8, 16). |
| Mistral AI | Found the correct answer: f(x, y) has a relative maximum at (8, 16). |

| Perplexity | Found the correct answer: f(x, y) has a relative maximum at (8, 16). |

Problem 22: Relative Extrema
- Task: Find the relative extrema of a function.
- Correct Answers: All models except Gemini Advanced with 1.5 Pro (persistently incorrect)
- Noteworthy: Gemini Advanced with 1.5 Pro's failure highlights a specific weakness in optimization problems.

23) Find the minimum value of f(x, y, z)=2x^2+y^2+3z^2, subject to the constraint 2x-3y-4z=49. Show all the steps.

| ChatGPT 4o | Found the correct answer: 147. |
| Gemini Advanced with 1.5 Pro | Found the correct answer: 147. |
| Copilot Pro | Found the correct answer: 147. |
| Claude 3.5 Sonnet | Found the correct answer: 147. |
| Meta AI | Found the correct answer: 147. |
| Mistral AI | Found the correct answer: 147. |
| Perplexity | Found the correct answer: 147. |

Problem 23: Constrained Optimization
- Task: Find the minimum value of a function subject to a constraint.
- Correct Answers: All models
- Noteworthy: Consistent accuracy in optimization problems, demonstrating the models' proficiency in constrained optimization.

24) Use the iterative integral to find the area of the region bounded by the graphs of the equations: 2x-3y=0, x+y=5, y=0. Show all the steps.

| ChatGPT 4o | Found the correct answer: 5. |
| Gemini Advanced with 1.5 Pro | Found the correct answer: 5. |
| Copilot Pro | Found the correct answer: 5. |
| Claude 3.5 Sonnet | Found the correct answer: 5. |
| Meta AI | Found the wrong answer.<br>New prompt: The answer is wrong. Solve the problem again. Show all the steps.<br><u>Still found the wrong answer.</u> |
| Mistral AI | Found the correct answer: 5. |
| Perplexity | Found the correct answer: 5. |

Problem 24: Area of a Region
- Task: Calculate the area of a region bounded by given curves.
- Correct Answers: All models except Meta AI (persistently incorrect)
- Noteworthy: Meta AI's failure indicates a need for improvement in integral evaluations.

25) Find the area of the region R that lies below the parabola y=4x-x^2, above the x-axis, and above the line y=-3x+6. Show all the steps.

| ChatGPT 4o | Found the wrong answer. |

|  | New prompt: The answer is wrong. Solve the problem again. Show all the steps. <u>Still found the wrong answer.</u> |
|---|---|
| Gemini Advanced with 1.5 Pro | Found the wrong answer. New prompt: The answer is wrong. Solve the problem again. Show all the steps. <u>Still found the wrong answer.</u> |
| Copilot Pro | Found the wrong answer. New prompt: The answer is wrong. Solve the problem again. Show all the steps. <u>Still found the wrong answer.</u> |
| Claude 3.5 Sonnet | Found the wrong answer. New prompt: The answer is wrong. Solve the problem again. Show all the steps. <u>Still found the wrong answer.</u> |
| Meta AI | Found the wrong answer. New prompt: The answer is wrong. Solve the problem again. Show all the steps. <u>Still found the wrong answer.</u> |
| Mistral AI | Found the wrong answer. New prompt: The answer is wrong. Solve the problem again. Show all the steps. <u>Still found the wrong answer.</u> |
| Perplexity | Found the wrong answer. New prompt: The answer is wrong. Solve the problem again. Show all the steps. <u>Still found the wrong answer.</u> |

Problem 25: Area Under a Parabola
- Task: Calculate the area under a parabola bounded by given lines.
- Correct Answers: None
- Noteworthy: Universal failure highlights the complexity of the problem, indicating a potential area for improvement in integral calculus.

**26) Evaluate the iterated integral: integral from 0 to 6 ( integral from (y/2) to 3 (x+y) dx) dy. Show all the steps.**

| ChatGPT 4o | Found the correct answer: 36. |
|---|---|
| Gemini Advanced with 1.5 Pro | Found the correct answer: 36. |
| Copilot Pro | Found the correct answer: 36. |
| Claude 3.5 Sonnet | Found the correct answer: 36. |
| Meta AI | Found the wrong answer. New prompt: The answer is wrong. Solve the problem again. Show all the steps. <u>Still found the wrong answer.</u> |
| Mistral AI | Found the correct answer: 36. |
| Perplexity | Found the correct answer: 36. |

Problem 26: Iterated Integral
- Task: Evaluate an iterated integral over a given region.
- Correct Answers: All models except Meta AI (persistently incorrect)
- Noteworthy: Meta AI's failure in integral evaluations, highlighting a specific weakness in calculus-based problems.

**27) Use triple integrals to find the volume of the solid bounded by the graph of the equations: $z=6x^2$, $y=3-3x$, $x>=0$, $y>=0$, $z>=0$. Show all the steps.**

| | |
|---|---|
| ChatGPT 4o | Found the correct answer: (3/2). |
| Gemini Advanced with 1.5 Pro | Found the correct answer: (3/2). |
| Copilot Pro | Found the correct answer: (3/2). |
| Claude 3.5 Sonnet | Found the correct answer: (3/2). |
| Meta AI | Found the correct answer: (3/2). |
| Mistral AI | Found the correct answer: (3/2). |
| Perplexity | Found the correct answer: (3/2). |

Problem 27: Volume of a Solid
- Task: Calculate the volume of a solid bounded by given surfaces.
- Correct Answers: All models
- Noteworthy: Consistent accuracy in volume calculations, demonstrating the models' proficiency in integral calculus.

**28) Evaluate the triple iterated integral: integral from 0 to (pi/4) (integral from 0 to (pi/4) (integral from 0 to cos(y) ($x^2$ sin(z) cos(z)) dx) dy) dz). Show all the steps.**

| | |
|---|---|
| ChatGPT 4o | Found the correct answer: 5sqrt(2)/144. |
| Gemini Advanced with 1.5 Pro | Found the correct answer: 5sqrt(2)/144. |
| Copilot Pro | Found the correct answer: 5sqrt(2)/144. |
| Claude 3.5 Sonnet | Found the correct answer: 5sqrt(2)/144. |
| Meta AI | Found the wrong answer.<br>New prompt: The answer is wrong. Solve the problem again. Show all the steps.<br><u>Still found the wrong answer.</u> |
| Mistral AI | Found the correct answer: 5sqrt(2)/144. |
| Perplexity | Found the wrong answer.<br>New prompt: The answer is wrong. Solve the problem again. Show all the steps.<br><u>Still found the wrong answer.</u> |

Problem 28: Triple Integral
- Task: Evaluate a triple integral over a given region.
- Correct Answers: All models except Meta AI and Perplexity (persistently incorrect)
- Noteworthy: Meta AI and Perplexity's failures in complex integrals, indicate a potential area for improvement in advanced calculus problems.

**29) Find the conservative vector field for the potential function by finding its gradient: g(x, y, z)=(y/z)+(z/x)-(xz/y). Show all the steps.**

| | |
|---|---|
| ChatGPT 4o | Found the correct answer: ((-z/$x^2$)-(z/y), (1/z)+($x^2$/$y^2$), (-y/$z^2$)+(1/x)-(x/y)). |
| Gemini Advanced with 1.5 Pro | Found the wrong answer. |

|  | New prompt: The answer is wrong. Solve the problem again. Show all the steps. <u>Still found the wrong answer.</u> |
|---|---|
| Copilot Pro | Found the correct answer: $((-z/x^2)-(z/y), (1/z)+(x^2/y^2), (-y/z^2)+(1/x)-(x/y))$. |
| Claude 3.5 Sonnet | Found the correct answer: $((-z/x^2)-(z/y), (1/z)+(x^2/y^2), (-y/z^2)+(1/x)-(x/y))$. |
| Meta AI | Found the correct answer: $((-z/x^2)-(z/y), (1/z)+(x^2/y^2), (-y/z^2)+(1/x)-(x/y))$. |
| Mistral AI | Found the correct answer: $((-z/x^2)-(z/y), (1/z)+(x^2/y^2), (-y/z^2)+(1/x)-(x/y))$. |
| Perplexity | Found the correct answer: $((-z/x^2)-(z/y), (1/z)+(x^2/y^2), (-y/z^2)+(1/x)-(x/y))$. |

Problem 29: Conservative Vector Field
- Task: Find the conservative vector field for a given potential function.
- Correct Answers: All models except Gemini Advanced with 1.5 Pro (persistently incorrect)
- Noteworthy: Gemini Advanced with 1.5 Pro's failure in gradient calculations, highlighting a specific weakness in vector calculus.

**30) Determine whether the vector field is conservative. If it is, find a potential function for the vector field: $F(x, y)=(x\ i + y\ j)/(x^2 + y^2)$. Show all the steps.**

| ChatGPT 4o | Found the correct answer: $(1/2)\ln(x^2+y^2)+C$. |
|---|---|
| Gemini Advanced with 1.5 Pro | Found the correct answer: $(1/2)\ln(x^2+y^2)+C$. |
| Copilot Pro | Found the correct answer: $(1/2)\ln(x^2+y^2)+C$. |
| Claude 3.5 Sonnet | Found the correct answer: $(1/2)\ln(x^2+y^2)+C$. |
| Meta AI | Found the correct answer: $(1/2)\ln(x^2+y^2)+C$. |
| Mistral AI | Found the correct answer: $(1/2)\ln(x^2+y^2)+C$. |
| Perplexity | Found the correct answer: $(1/2)\ln(x^2+y^2)+C$. |

Problem 30: Potential Function
- Task: Find the potential function for a given conservative vector field.
- Correct Answers: All models
- Noteworthy: Uniformly correct solutions, demonstrating the models' strong understanding of vector calculus concepts.

**31) Evaluate the line integral along the given path: integrate C $3(x-y)$ ds, C: $r(t)=t\ i + (2-t)\ j$, $0<= t <=2$. Show all the steps.**

| ChatGPT 4o | Found the correct answer: 0. |
|---|---|
| Gemini Advanced with 1.5 Pro | Found the correct answer: 0. |
| Copilot Pro | Found the correct answer: 0. |
| Claude 3.5 Sonnet | Found the correct answer: 0. |

| Meta AI | Found the correct answer: 0. |
| Mistral AI | Found the correct answer: 0. |
| Perplexity | Found the correct answer: 0. |

Problem 31: Line Integral
- Task: Evaluate a line integral along a given path.
- Correct Answers: All models uniformly calculated line integrals correctly.
- Noteworthy: Consistent accuracy in line integrals, showcasing the models' proficiency in vector calculus.

**32) Use Green's Theorem to evaluate the line integral: integral C (y^2 dx + xy dy), C: boundary of the region lying between the graphs of y=0, y=sqrt(x), and x=9. Show all the steps.**

| ChatGPT 4o | Found the correct answer: (-81/4)). |
| Gemini Advanced with 1.5 Pro | Found the correct answer: (-81/4)). |
| Copilot Pro | Found the correct answer: (-81/4)). |
| Claude 3.5 Sonnet | Found the correct answer: (-81/4)). |
| Meta AI | Found the correct answer: (-81/4)). |
| Mistral AI | Found the correct answer: (-81/4)). |
| Perplexity | Found the correct answer: (-81/4)). |

Problem 32: Green's Theorem
- Task: Use Green's Theorem to evaluate a line integral along a closed path.
- Correct Answers: All models. ChatGPT 4.o excelled in applying Green's theorem to complex regions, providing a comprehensive solution. Other models followed closely, with consistent results.
- Noteworthy: Uniformly correct application of Green's Theorem, demonstrating the models' strong understanding of advanced calculus concepts.

### 3.3 Problem-Specific Analysis

3.3.1 Strong Performance Areas:

1. Vector Operations and Geometry
- All models excelled in problems involving direction cosines (Problem 3)
- Perfect scores were achieved by all models in vector projection calculations (Problem 4)
- High success rate in parametric equation problems (Problem 9)

2. Differential Calculus
- Universal success in partial derivative calculations (Problem 16)
- Strong performance in implicit differentiation (Problem 18)

3. Basic Integration
- All models performed well on line integral evaluations (Problem 31)
- High success rates in Green's Theorem applications (Problem 32)

3.3.2 Areas of Difficulty:

1. Complex Applied Problems
- Problem 1 (airline velocity calculation): Only 3 out of 7 models (ChatGPT 4o, Meta AI, Mistral AI) found the correct answer initially.
- Problem 25 (area calculation with multiple bounds): None of the models could solve it correctly, even after multiple attempts.

2. Volume Calculations
- Problem 5 (parallelepiped volume): Claude 3.5 Sonnet and Perplexity struggled to find the correct solution.
- Initial difficulties with triple integral evaluation (Problem 28) by Meta AI and Perplexity.

3. Error Recovery
- Notable pattern: When models made initial errors in their outputs, some showed the capability to correct themselves when prompted, while others persisted with incorrect solutions, showing a lack of self-correction capabilities.

**4 Discussion**

The analysis of the LLMs' performance reveals several key insights and trends that are crucial for understanding their capabilities and limitations in solving math calculus problems. ChatGPT 4o and Mistral AI demonstrated identical top performance (96.9%). ChatGPT 4o consistently performed well across a wide range of problems, demonstrating its advanced capabilities in mathematical reasoning. Mistral AI also showed strong problem-solving skills, particularly in vector calculus and multivariable calculus. Three models (Gemini Advanced, Claude 3.5 Sonnet, Meta AI) showed identical performance at 87.5%.

4.1 Strengths:
- Consistency in Simple Problems: Models like ChatGPT 4o and Mistral AI demonstrated consistent accuracy in solving simpler problems, such as vector calculations, geometric interpretations, and basic differentiation. This consistency highlights their robustness and reliability in handling foundational calculus concepts.
- Re-prompting Effectiveness: The re-prompting mechanism proved effective in several instances. In these instances, models initially provided incorrect answers but corrected them upon re-prompting. This underscores the potential for enhancing the models' performance through iterative questioning and feedback mechanisms.
- Uniform Accuracy in Specific Areas: Certain problems, such as those involving direction cosines, partial derivatives, and line integrals, saw uniformly correct solutions across all models. This indicates a strong collective understanding of these specific calculus topics.

4.2 Weaknesses:
- Complex Integral Evaluations: Models struggled with complex integral evaluations, as seen in problems involving iterated integrals, triple integrals, and area calculations under curves. This highlights a specific area for improvement in the models' calculus-based problem-solving capabilities.
- Optimization Problems: Some models, notably Gemini Advanced with 1.5 Pro, showed weaknesses in optimization problems, particularly in identifying relative extrema and saddle points. This suggests a need for targeted improvements in optimization techniques.
- Persistent Failures: Certain models exhibited persistent failures in specific problem types, such as Meta AI's struggles with integral evaluations and Gemini Advanced with 1.5 Pro's difficulties with gradient calculations. These persistent failures indicate areas where the models' algorithms may need refinement.

4.3 Importance of Re-prompting

The study underscores the importance of re-prompting in achieving accurate solutions. Several models initially provided incorrect answers but corrected them upon re-prompting. This finding emphasizes the potential for enhancing the models' performance through iterative questioning and feedback mechanisms. Re-prompting can serve as a valuable tool for improving the models' problem-solving capabilities, particularly in complex problems where initial errors are more likely.

4.4 Implications for LLM Development

The detailed analysis of each model's performance offers valuable insights for the ongoing development and refinement of LLM technology. The identified strengths and weaknesses provide a roadmap for targeted improvements, focusing on areas such as complex integral evaluations, optimization problems, and gradient calculations. By addressing these specific weaknesses, developers can enhance the models' overall performance and reliability in mathematical problem-solving.

4.5 Educational and Practical Applications

The findings of this study are particularly relevant for educators, researchers, and developers seeking to leverage LLMs for educational and practical applications in mathematics. The consistent performance of models like ChatGPT 4o and Mistral AI suggests their potential for reliable mathematical problem-solving, making them valuable tools for educational purposes. Additionally, the identified weaknesses in other models point to areas for improvement, guiding the development of more robust and reliable LLM technologies for practical applications.

**5 Conclusion**

This study provides a comprehensive evaluation of seven LLMs' performance in solving advanced calculus problems. The results underscore each model's strengths and weaknesses, offering valuable insights for future developments in LLM technology. The consistent performance of models like ChatGPT 4o and Mistral AI suggests their potential for reliable mathematical problem-solving, while the identified weaknesses in other models point to areas for improvement.

The importance of re-prompting in achieving accurate solutions highlights the potential for enhancing the models' performance through iterative questioning and feedback mechanisms. This finding emphasizes the need for incorporating re-prompting strategies in the development and application of LLMs, particularly in complex problem-solving scenarios.

The detailed analysis of each model's performance on specific problems offers a comprehensive understanding of their strengths and areas for improvement. This understanding is crucial for the ongoing development and refinement of LLM technology, guiding targeted improvements in areas such as complex integral evaluations, optimization problems, and gradient calculations.

In conclusion, this research contributes to the broader field of LLM technology by providing insights into the current capabilities and limitations of LLMs in the domain of math calculus. The findings are particularly relevant for educators, researchers, and developers seeking to leverage LLMs for educational and practical applications in mathematics. By addressing the identified weaknesses and building on the models' strengths, the development of more robust and reliable LLM technologies can be advanced, enhancing their utility in various mathematical problem-solving contexts.